\documentclass[10pt,twocolumn,letterpaper]{article}

\usepackage{cvpr}      

\usepackage{graphicx}
\usepackage{multirow}
\usepackage{amsmath}
\DeclareMathOperator*{\argmax}{arg\,max}

\usepackage{amssymb}
\usepackage{booktabs}
\usepackage{mathtools}

\usepackage{bbm}

\usepackage[pagebackref,breaklinks,colorlinks]{hyperref}

\usepackage[capitalize]{cleveref}
\crefname{section}{Sec.}{Secs.}
\Crefname{section}{Section}{Sections}
\Crefname{table}{Table}{Tables}
\crefname{table}{Tab.}{Tabs.}

\usepackage{color}

\def\cvprPaperID{4154} 
\def\confName{CVPR}
\def\confYear{2022}

\begin{document}

\title{Point2Seq: Detecting 3D Objects as Sequences}

\author{
Yujing Xue\textsuperscript{1,2}\thanks{
Joint first authors with equal contribution.
} 
\quad Jiageng Mao\textsuperscript{3}\footnotemark[1]
\quad Minzhe Niu\textsuperscript{2}
\quad Hang Xu\textsuperscript{2}
\quad Michael Bi Mi\textsuperscript{4}\\
\quad Wei Zhang\textsuperscript{2}
\quad Xiaogang Wang\textsuperscript{3}
\quad Xinchao Wang\textsuperscript{1}\thanks{Corresponding author. Email: {\tt xinchao@nus.edu.sg}}
\\
\textsuperscript{1}{National University of Singapore}
\quad \textsuperscript{2}{Huawei Noah's Ark Lab}\\
\quad \textsuperscript{3}{The Chinese University of Hong Kong}
\quad \textsuperscript{4}{Huawei International Pte Ltd}
}

\maketitle

\begin{abstract}
We present a simple and effective framework, 
named Point2Seq, 
for 3D object detection from point clouds. 
In contrast to previous methods that normally 
{predict attributes of 3D objects all at once}, 
we expressively model the 
interdependencies between 
attributes of 3D objects,
which in turn enables
a better detection accuracy. 
Specifically, we view each 3D object as a sequence of words
and reformulate the 3D object detection task as 
decoding words from 3D scenes in an auto-regressive manner.
We further propose a lightweight scene-to-sequence decoder
that can auto-regressively generate words conditioned on 
features from a 3D scene as well as cues from the preceding words. 
The predicted words eventually constitute a set of sequences 
that completely describe the 3D objects in the scene, 
and all the predicted sequences are then automatically 
assigned to the respective ground truths
through similarity-based sequence matching.
Our approach is conceptually intuitive
and can be readily plugged upon most existing 3D-detection
backbones without adding too much computational overhead;
the sequential decoding paradigm we proposed, 
on the other hand,
can better exploit information from  complex 3D scenes 
with the aid of preceding predicted words.
Without bells and whistles, 
our method significantly outperforms previous 
anchor- and center-based 3D object detection frameworks,
yielding the new state of the art on the
challenging ONCE dataset as well as the Waymo Open Dataset.
Code is available at \url{https://github.com/ocNflag/point2seq}. 

\end{abstract}



\section{Introduction}

\begin{figure}[ht]
  \begin{center}
    \includegraphics[width=0.46\textwidth]{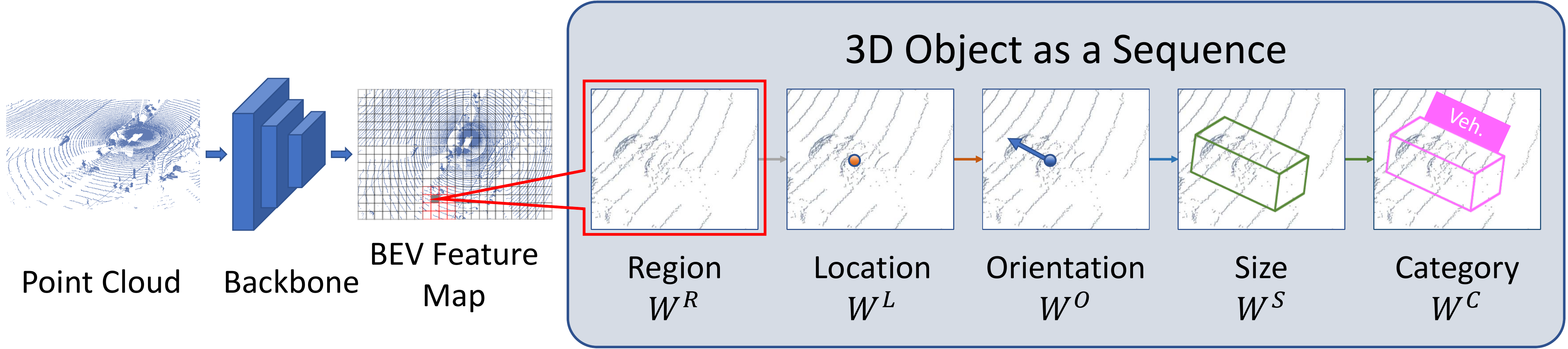}
  \end{center}
  \setlength{\abovecaptionskip}{0.cm}
  \caption{Point2Seq reformulates the 3D object detection problem as auto-regressively generating sequences of words that can represent the 3D objects. The sequential decoding paradigm attains better detection performance compared to previous works that predict all attributes of the 3D objects in parallel,
  thanks to its competence in exploring intrinsic dependencies among object attributes.
  }
  \label{fig:intro}
\vspace{-6mm}
\end{figure}

3D object detection is a critical component of intelligent perception systems for self-driving, aiming to localize and recognize cars, pedestrians, and other key objects around an autonomous vehicle. With the increasing popularity of LiDAR sensors, 3D object detection from point clouds has been receiving much attention owing to the advanced detection accuracy compared to other input modalities. 


In driving scenes, most 3D objects are extremely small relative to the detection range, and numerous approaches have been employed to detect those small objects from point clouds accurately. Anchor-based methods~\cite{yan2018second,lang2019pointpillars} place predefined anchors on each pixel center of a Bird-Eye-View (BEV) feature map, while center-based approaches~\cite{yin2021center, ge2020afdet} treat the pixels near object centers as positives and predict boxes using those pixel features. 
These methods rely on complex hand-crafted procedures of label assignments and post-processing, and quantization errors introduced by the BEV representations lead to severe misalignment between the object locations and the pixel features used to predict those objects. Approaches like~\cite{shi2020pv, shi2020points, mao2021pyramid} rely on a second refinement stage to mitigate the misalignment issue, yet at the cost of adding too much computational overhead. 
Therefore, learning more spatially-aligned features to detect the 3D objects accurately while maintaining a high efficiency poses an open challenge to the research community.


To address the challenge, in this paper, we introduce Point2Seq, a flexible and streamlined framework for 3D object detection from point clouds. 
Unlike prior methods that typically predict all attributes of a 3D object (\eg, location, class, size) simultaneously, we represent each object as a sequence, in which each word corresponds to an object attribute, and we explicitly explore the inherent dependencies among words alongside their relations with the input 3D scene to progressively predict each attribute of 3D objects.
Our motivation is quite intuitive: given the fact that each object is represented by sequential words, the existing words will provide 3D detectors with cues to better exploit spatial features and help detectors predict the following words more accurately. For instance, 3D detectors can leverage more spatially-aligned features for an object if the object location has been formerly predicted and better recognize the object class if its size information has already been known. 
It is therefore desirable to design a detection framework that may
sequentially predict words of 3D objects conditioned
on the preceding generated words as well as the spatial features until all the words form a set of sequences describing the 3D objects in the scene. 

To achieve this goal, we must address two critical challenges: how to design the sequential object words prediction module and make it compatible with the existing 3D detection pipelines and how to optimize the 3D detector with the ground truth and predicted sequences. To resolve the first problem, we propose a novel scene-to-sequence decoder, which takes the BEV feature map and a set of initial region cues as input and auto-regressively decodes sequences for all objects in parallel. The scene-to-sequence decoder is compatible with most grid-based 3D backbones~\cite{yan2018second, zhou2018voxelnet, wang2020pillar, lang2019pointpillars, mao2021voxel} and can effectively aggregate features from those backbones based on the information of the preceding words. By virtue of the highly-parallel deep learning libraries, the scene-to-sequence decoder can generate the sequences of all 3D objects in one shot, with little added time and memory cost. 

To handle the second issue, we adopt the set-to-set loss to match the predicted sequences with the ground truths. Unlike existing approaches~\cite{wang2021object, misra2021end, carion2020end} that utilize the sum of the classification and regression loss as the cost function for bipartite matching, in this paper, we propose a novel metric to measure the similarity between two sequences. Then we perform bipartite matching by maximizing the global similarity of the prediction and the ground truth set using the proposed metric. In this manner, each predicted sequence can be automatically assigned to a respective ground truth without pre-defined anchors or centers. The assignments are globally optimal and result in better performance compared to previous methods.



With the lightweight scene-to-sequence decoder, our method can progressively predict the words of 3D objects, yielding reliable predictions that significantly outperform state-of-the-art.
In addition, our method is free from the human-designed procedures of label assignments with similarity-based sequence matching.
Our key contributions are summarized as follows:

$\bullet$ We present an effective and flexible framework for 3D object detection from point clouds. We represent each 3D object as a sequence of words and model the 3D object detection problem as decoding the words from the 3D scenes in an auto-regressive manner.

$\bullet$ We propose a scene-to-sequence decoder that can auto-regressively generate sequences representing the detected 3D objects and introduce the similarity-based sequence matching scheme to enable automatic assignments of the predicted sequences to the respective ground truths for end-to-end training.

$\bullet$ Our method significantly outperforms the anchor-based and center-based 3D detectors with the same backbone, attaining $66.16\%$ mAP on the ONCE dataset and $77.52\%$ vehicle L1 mAP on the Waymo Open Dataset.

\section{Related Work}

\textbf{Backbones for 3D object detection.} 3D detectors rely on various backbone networks to extract features from input point clouds. The existing backbones of 3D detectors can be divided into $3$ streams: point-based, range-based, and grid-based. The point-based backbones~\cite{shi2019pointrcnn,Yang2020Factor,shi2020point,yang20203dssd,Yang2020Distill,LiuHuihuiAAAI20,pan20213d} operate directly on raw point clouds with the point operators~\cite{qi2017pointnet, mao2019interpolated} to extract the point-wise features. The range-based backbones~\cite{sun2021rsn, liang2020rangercnn, Meyer2019LaserNetAE, fan2021rangedet, bewley2020range} take range images, which are the raw data from LiDAR sensors, as the input representation. Customized operators, \eg, range conditional convolutions~\cite{bewley2020range}, meta-kernels~\cite{fan2021rangedet}, are applied on the range images for feature extraction. The grid-based backbones~\cite{lang2019pointpillars, wang2020pillar, yan2018second, zhou2018voxelnet, mao2021voxel} firstly rasterize point clouds into voxels or pillars. Those voxels or pillars are fed into a 3D network and then projected into a BEV feature map, followed by a 2D convolutional neural network to detect 3D objects. Among the $3$ kinds of backbones, the grid-based backbones can obtain superior detection performance while maintaining high efficiency. Our Point2Seq is a flexible detection framework and can be applied to most grid-based backbones.   

\begin{figure*}[ht]
  \begin{center}
    \includegraphics[width=1\textwidth]{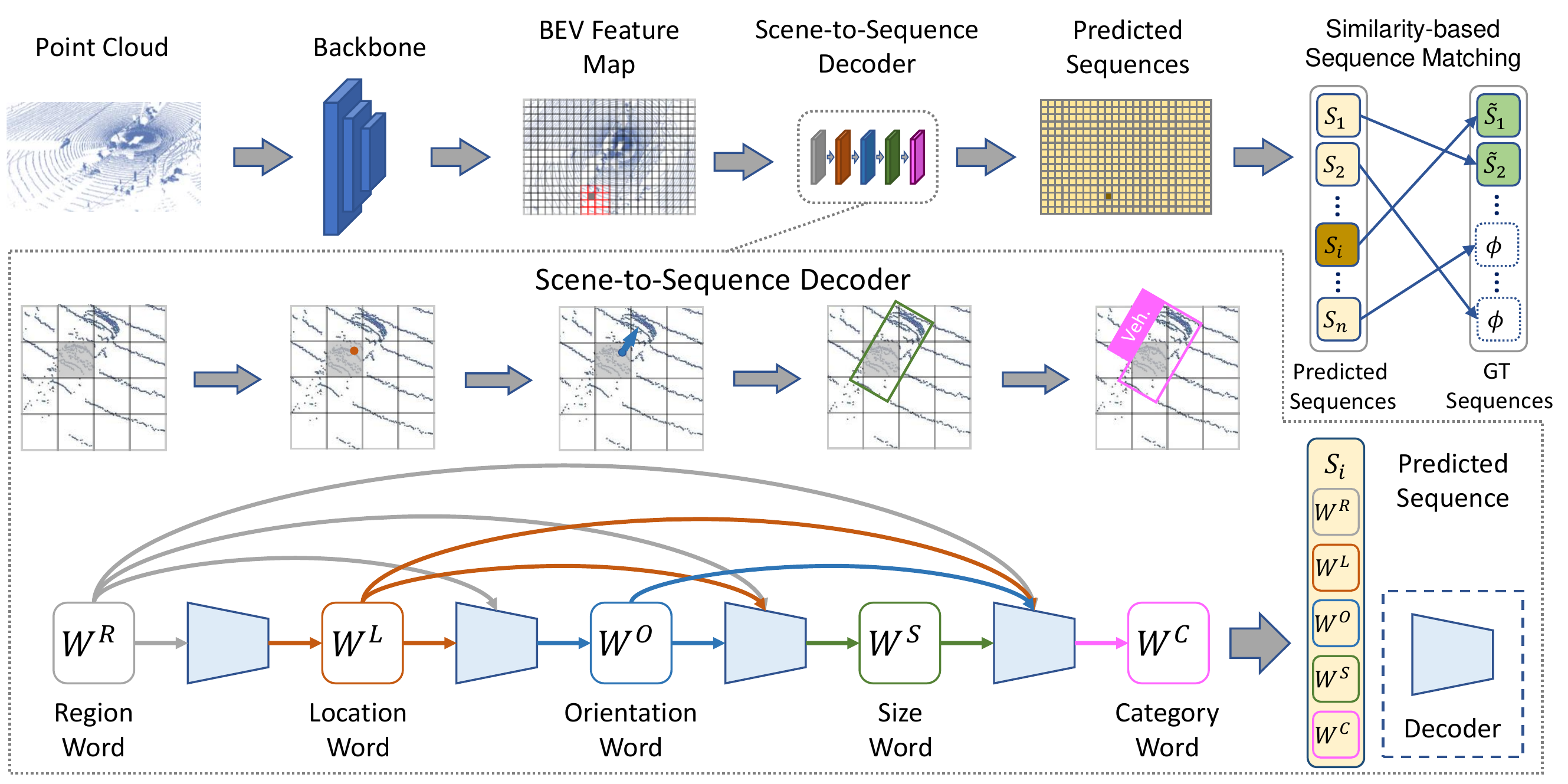}
  \end{center}
  \setlength{\abovecaptionskip}{0.cm}
  \caption{The overall architecture of our framework. Point2Seq contains three major components: the 3D backbone, the scene-to-sequence decoder, and the similarity-based sequence matching scheme. The 3D backbone takes the rasterized point cloud as input and outputs the Bird-Eye-View (BEV) feature map for the 3D scene. The scene-to-sequence decoder operates on the BEV feature map and sequentially predicts the words of 3D objects based on the information from the preceding predicted words. Finally, the predicted sequences are automatically assigned to the corresponding ground truths by the proposed similarity-based sequence matching scheme.}
  \label{fig:method}
\vspace{-5mm}
\end{figure*}

\textbf{3D objects prediction mechanisms.} 3D detectors adopt various prediction mechanisms to generate detected 3D objects from the backbone features. For the point-based backbones, PointRCNN~\cite{shi2019pointrcnn} directly generates object proposals on the key points' locations. For the range-based backbones, RangeDet~\cite{fan2021rangedet} generates 3D bounding boxes on the pixels of range images. For the grid-based backbones, SECOND~\cite{yan2018second} places a set of 3D anchors on every grid center of a BEV map. The anchors that have a high overlap with the ground truth 3D objects are set to positives, and the objects are then predicted on the positive anchors. SA-SSD~\cite{he2020structure} applies the part-sensitive warping scheme to enhance the spatial features. CenterPoints~\cite{yin2021center} treats BEV pixels near the object centers as positives and generates predicted bounding boxes near the object centers. Existing methods usually predict all attributes of the 3D objects simultaneously without considering the intra-object information, while Point2Seq can model the interdependencies among the object's attributes with the scene-to-sequence decoder.

\textbf{Set-to-set matching for object detection.} The set-to-set matching mechanism is first introduced in DETR~\cite{carion2020end} in which a set of object queries are assigned to the respective ground truths through bipartite matching. This method is improved by a series of works~\cite{zhu2020deformable, sun2021sparse, sun2020onenet, wang2021end} on image-based object detection. The set-to-set matching scheme has also been adopted in 3D object detection. 3DETR~\cite{misra2021end} utilizes sampled points as object queries. Object DGCNN~\cite{wang2021object} leverages a sparse set of object queries to iteratively interact with a BEV feature map to generate 3D objects. Compared to previous methods, Point2Seq does not require the sparse object queries or the multi-step refinement modules in~\cite{vaswani2017attention, wang2019dynamic}. The sequences are generated densely from the BEV map in parallel and are automatically matched to the corresponding ground truths by similarity-based sequence matching, without additional modules or parameters. 

\section{Detecting 3D Objects as Sequences}

\subsection{Architecture} \label{3.1}
For each 3D scene, Point2Seq takes a point cloud as input and outputs a set of 3D bounding boxes $\mathcal{B} = \{B_{1}, \cdots, B_{M}\} \in \mathbb{R}^{M \times 8}$ that represent the detected 3D objects, \eg, vehicles, pedestrians, cyclists, \etc. A 3D point cloud is an $N \times d$ matrix, where $N$ denotes the number of points in the scene and $d$ denotes the initial features of points, \ie, 3D coordinates, intensity, \etc. Each 3D object $B_{i} \in \mathbb{R}^{8}$ is a vector: $[x, y, z, l, w, h, \theta, c]$, where $[x, y, z]$ is the location of the object's center, $[l, w, h]$ is the object's size, $\theta$ is the object's orientation, and $c$ is the class of the object.    

As is shown in Figure~\ref{fig:method}, the architecture of Point2Seq is composed of $3$ parts: the 3D backbone, the scene-to-sequence decoder, and the similarity-based sequence matching scheme. The 3D backbone first consumes a point cloud and generates a BEV feature map from the point cloud. Then the scene-to-sequence decoder takes both the BEV feature map and an initial set of region cues as input and decodes sequences of words that describe the detected 3D objects in the scene. Finally, the similarity-based sequence matching is applied to assign the predicted sentences to the respective ground truths. The choice of 3D backbones in Point2Seq can be flexible: most grid-based backbones~\cite{yan2018second, lang2019pointpillars, wang2020pillar, mao2021voxel} can be applied in our framework. The grid-based backbones first transform point clouds into voxels or pillars, and then 3D features are extracted from those voxels or pillars by sparse convolutions~\cite{graham20183d} or set abstraction~\cite{qi2017pointnet}, respectively. The 3D features are then projected to Bird-Eye-View (BEV). A 2D convolutional neural network is applied on the projected features to obtain the final BEV feature map $F \in \mathbb{R}^{H \times W \times C}$, where the detection space is divided into an $H \times W$ grid, and $C$ denotes the number of feature channels. 

In the scene-to-sequence decoder, the 3D objects $\mathcal{B}$ are transformed into a set of sequences $\{S_{1}, \cdots, S_{M}\}$, where each sequence $S_{i}$ corresponds to a 3D object $B_{i}$ and contains $K$ words $\{W_{i}^{0}, \cdots, W_{i}^{K-1}\}$ that can represent the 3D object. The scene-to-sequence decoder operates on the BEV feature map $F$ and can auto-regressively predict a word $W_{i}^{j}$ conditioning on $F$ and the preceding words $W_{i}^{0:j-1}$. We will introduce the detailed design of the scene-to-sequence decoder in Sec.~\ref{3.2} and then illustrate how to optimize the scene-to-sequence decoder and the 3D backbone through similarity-based sequence matching in Sec.~\ref{3.3}. Finally, we discuss and compare our method with previous literature in Sec.~\ref{3.4}.   

\subsection{Scene-to-Sequence Decoder} \label{3.2}

\textbf{Problem formulation.} Previous single-stage 3D detectors model the 3D object detection problem as predicting all attributes of the 3D objects $\mathcal{B}$ simultaneously from the features $F$. The learning process can be formulated as the optimization problem:
\begin{equation}
    \max \sum_{i=1}^{M} log P(B_{i} | \mathcal{D}(F)),
\end{equation}
where $B_{i} \in \mathcal{B}$ is the attributes $[x, y, z, l, w, h, \theta, c]$ of the $i$th ground truth 3D object, $M=|\mathcal{B}|$, and $\mathcal{D}$ is normally a convolutional prediction head applied on the BEV feature map $F$. The parallel prediction paradigm is widely adopted for its efficiency. However, it suffers from the misalignment between the object's actual location and the BEV features used for prediction due to the high quantization errors introduced by rasterization. Multi-stage refinements can be employed to mitigate the misalignment problem but will consequently introduce too much computational overhead.

In this paper, we draw inspirations from Language Modeling (LM)~\cite{brown2020language, devlin2018bert} in natural language processing applications and translate each 3D object $B$ into a sequence $S$ containing $K$ words $\{W^{0}, \cdots, W^{K-1}\}$:
\begin{equation}
    B = \mathcal{T}(S) = \mathcal{T}(W^{0}, \cdots, W^{K-1}).
\end{equation}
The translation $\mathcal{T}$ is parameter-free and bidirectional, so the 3D objects and their corresponding words can be easily transformed into each other. Then similar to the language models, we can reformulate the detection problem as maximizing the probability production of all target words $\tilde{W_{i}}^{j}$ conditioning on the feature $F$ and the preceding predicted words $W_{i}^{0:j-1}$:
\begin{equation}
    \max \sum_{i=1}^{M} \sum_{j=1}^{K-1} log P(\tilde{W_{i}}^{j} | \mathcal{D}(F, W_{i}^{0}, \cdots, W_{i}^{j-1})).
\end{equation}
The main insight of our approach is that each 3D object is decomposed into several words, and predicting these words sequentially, instead of simultaneously in previous methods, enables more effective exploitation of the BEV features with the cues from the preceding predicted words.  

\textbf{3D object as a sequence of words.} Translating every 3D object into words is a pivotal step in our method. Different from those methods that adopt discrete tokens as words in natural language processing tasks, we represent the words in a continuous format in our method. The use of continuous representations for object words is preferable in 3D object detection since most attributes of 3D objects, \eg, locations, sizes, orientations, are continuous values, and the predicted words can be directly transformed back into the corresponding object's attributes without loss of accuracy. 

In this paper, each 3D object $B = [x, y, z, l, w, h, \theta, c]$ is translated into $5$ words: 
\begin{equation}
B = \mathcal{T}(S) = \mathcal{T}(W^{R}, W^{L}, W^{O}, W^{S}, W^{C}).  
\end{equation}
The region word $W^{R} = [R_{x}, R_{y}] \in \mathbb{R}^{2}$ indicates the possible region in which the 3D object is likely to appear, where $[R_{x}, R_{y}]$ is the BEV center coordinate of the region, and additional parameters $[R_{l}, R_{w}]$ are introduced to describe the spatial range of the region on the BEV feature map. The location word $W^{L} = [L_{x}, L_{y}, z] \in \mathbb{R}^{3}$ denotes the location of the object's center, where $L_{x} = (x - R_{x})/R_{l}$ and $L_{y} = (y - R_{y})/R_{w}$ denote the relative location in the region. The orientation word $W^{O} = [sin(\theta), cos(\theta)] \in \mathbb{R}^{2}$ encodes the object's orientation $\theta$ by trigonometric functions. The size word $W^{S} = [log(l), log(w), log(h)] \in \mathbb{R}^{3}$ applies logarithmic functions to the object's size. The category word $W^{C} \in \mathbb{R}^{n+1}$ indicates the probabilities of $n$ detected classes and the background class. 

\textbf{Scene-to-sequence prediction.} Our proposed scene-to-sequence decoder takes the BEV features $F$ and a set of region words $W^{R}$ as the initial inputs and sequentially predicts $W^{L}$, $W^{O}$, $W^{S}$, $W^{C}$, \ie, $W^{1}$, $W^{2}$, $W^{3}$, $W^{4}$ in $4$ steps for each region $W^{R}$. In each step, the words are predicted on a hidden state map $H \in \mathbb{R}^{H \times W \times C}$ that encodes the historical information of the preceding steps. Specifically, the hidden state $H_{1}$ is firstly initialized as the input BEV feature map $F$:
\begin{equation}
    H_{1} = F.
\end{equation}

Then at the $j$th step, the word $W^{j}$ will be directly predicted from the hidden state $H_{j}$ at the corresponding region center $W^{R}$, \ie, $H_{j}[W^{R}] \in \mathbb{R}^{C}$, through a single linear projection layer $f_{linear}$:
\begin{equation}
    W^{j} = f_{linear}(H_{j}[W^{R}]),
\end{equation}
where $[\cdot]$ is the indexing operator. The hidden state $H_{j}[W^{R}]$ at $W^{R}$ will then be updated to the next step $H_{j+1}[W^{R}]$ based on the already learned knowledge from the former predicted words $\{W^{0}, \cdots, W^{j}\}$:
\begin{equation}
    H_{j+1}[W^{R}] = \Phi(H_{j}[W^{R}]; W^{0}, \cdots, W^{j}),
\end{equation}
where $\Phi$ is the update function. To model the hidden state update process at the $j$th step, near each region $W^{R}$, we first sample a sparse set of points $\{p^{j}_{1}, \cdots, p^{j}_{n}\} \in \mathbb{R}^{n \times 2}$ by a spatial sampler $\mathcal{S}$ parameterized by the predicted words:
\begin{equation}
\{p^{j}_{1}, \cdots, p^{j}_{n}\} = \mathcal{S}(W^{0}, \cdots, W^{j}).
\end{equation}
The sampling patterns of $\mathcal{S}$ are shown in Figure~\ref{fig:sampling}, and the detailed formulations are demonstrated in the appendix. Then for each region $W^{R}$, we can update to $H_{j+1}[W^{R}]$ by aggregating hidden vectors at those sampled locations on $H_{j}$, which can be formulated as
\begin{equation}
H_{j+1}[W^{R}] = \mathcal{A}(H_{j}[p^{j}_{1}], \cdots, H_{j}[p^{j}_{n}]),
\end{equation}
where the aggregation function $\mathcal{A}$ concatenates the sampled hidden vectors and projects them into the $\mathbb{R}^{C}$ space.

The initial set of the region words $W^{R}$ indicates where the 3D objects are likely to appear in the 3D scene. Since there is no such prior information, we employ a dense prediction strategy in this paper. Namely, we treat each pixel in the BEV feature map as a region word and predict a sequence for each BEV pixel. The category word $W^{C}$ is expected to predict the highest probability for the background class if there is no 3D object near the corresponding pixel region. The dense prediction paradigm benefits from the highly-parallel characteristics of the modern deep learning libraries, and the sequences for all pixels can be predicted in parallel by the scene-to-sequence decoder efficiently using shared MLPs and sampling operators. 

\begin{figure}[tp]
  \begin{center}
    \includegraphics[width=0.46\textwidth]{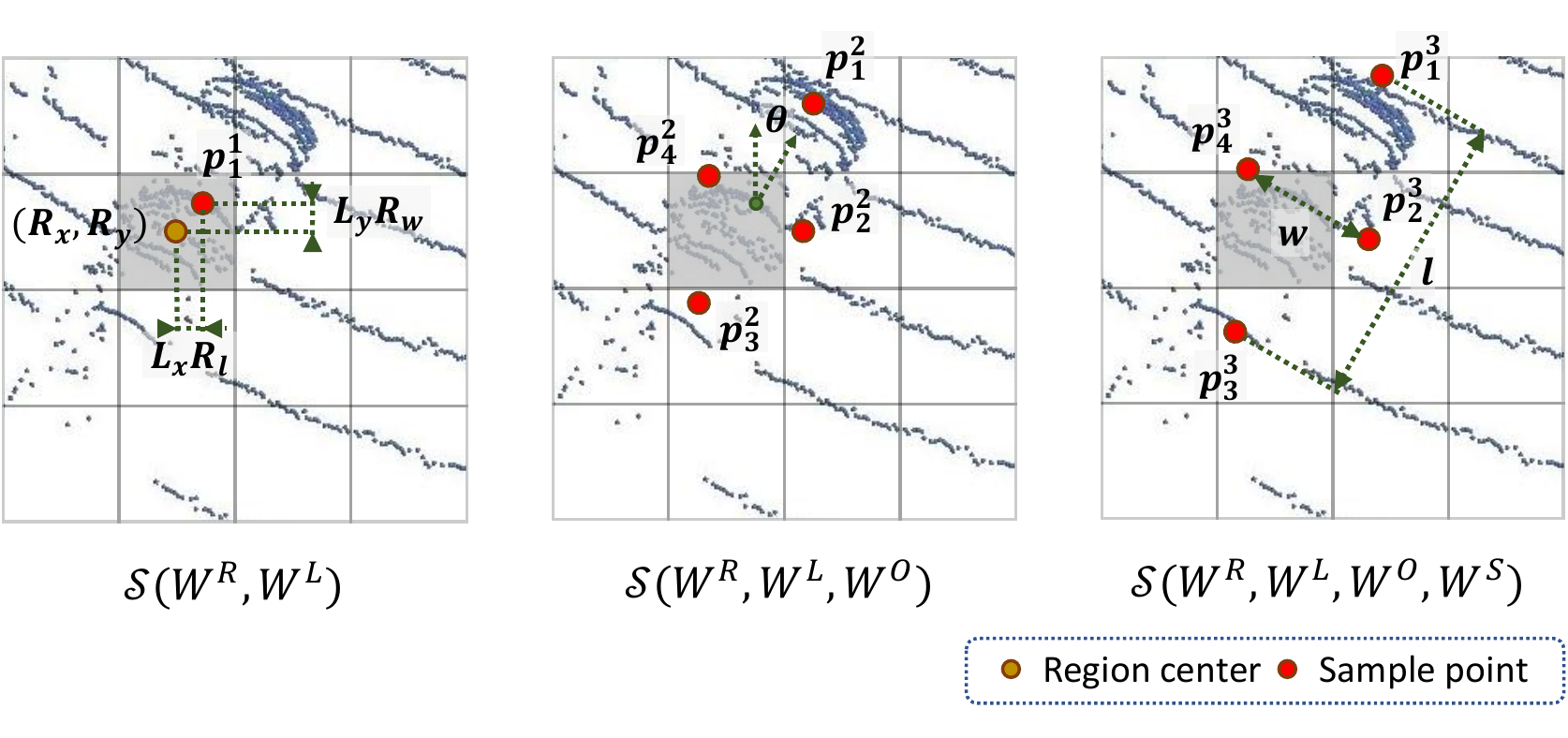}
  \end{center}
  \vspace{-6mm}
  \caption{The sample locations at each step. For each region, we progressively obtain $1$, $4$, $4$ points from the spatial sampler $\mathcal{S}$, based on the information from the predicted words.}
  \vspace{-4mm}
  \label{fig:sampling}
\end{figure}

\subsection{Similarity-based Sequence Matching} \label{3.3}
In this section, we will introduce how to optimize our detection framework by similarity-based sequence matching. Our approach is inspired by the set-to-set loss employed in the image-based detection frameworks~\cite{carion2020end}. Our main contribution lies in the design of a new cost function tailored for 3D object detection in the set-to-set matching problem. Specifically, we propose a novel metric $Sim(S, \tilde{S})$ to measure the similarity of two sequences $S$ and $\tilde{S}$ from the predicted and the ground truth sequence set, respectively. We match the predicted sequences to the corresponding target sequences by maximizing the global similarity of the two sequence sets. Finally, losses can be applied to the matched sequence pairs for back-propagation. In this manner, we can eliminate the hand-crafted label assignment process and make our model end-to-end trainable without non-maximum suppression.  

The scene-to-sequence decoder outputs a set of predicted sequences $\{S_{1}, \cdots, S_{HW}\}$ containing $H \times W$ sequences for all BEV pixels in total. For the ground truth 3D objects, we also construct a sequence set $\{\tilde{S}_{1}, \cdots, \tilde{S}_{M}, \varnothing, \cdots, \varnothing\}$ where the size of the ground truth set is equal to that of the prediction set, and we pad the remaining sequences with $\varnothing$ if $M < H \times W$. To measure the similarity between a predicted sequence $S$ and a ground truth sequence $\tilde{S}$, we define a new similarity metric that can be formulated as
\begin{equation} \label{eq_sim}
Sim(S, \tilde{S}) = (W^{C} \tilde{W}^{C})^{\alpha} \cdot e^{-(1-\alpha) \sum_{j \in \{R, L, O, S\}} |W^{j} - \tilde{W}^{j}| },
\end{equation}
where the first term $(W^{C} \tilde{W}^{C})^{\alpha}$ measures the class similarity between the predicted and the ground truth objects, and the second term measures the shape and location similarity. The hyper-parameter $\alpha$ is utilized to balance the two similarities and set to $0.25$ in our experiments. $Sim(S, \varnothing) = 0$ if a predicted sequence is matched to $\varnothing$. 

The proposed similarity metric is a more stringent criterion to match the predictions with the ground truths, and even slight differences between the two sequences can make the similarity score tend to $0$. Given the fact that the 3D objects are innately small and the mismatch should be completely avoided, the stringent similarity metric we proposed is preferable in the task of 3D object detection. 

With the similarity metric, we can further establish the optimal set-to-set matching $\Pi^{*}$ by considering the bipartite matching problem:
\begin{equation}
    \Pi^{*} = \argmax_{\Pi} \sum_{(i \rightarrow j) \in \Pi} Sim(S_{i}, \tilde{S}_{j}),
\end{equation}
where $\Pi$ is a bijective function that enables a one-to-one mapping from the predicted sequence set to the ground truth set. The bipartite matching problem aims to find the optimal $\Pi^{*}$ so that the maximum overall similarity of the two sets can be achieved. With $\Pi^{*}$, every ground truth sequence can be automatically assigned to the corresponding predicted sequence that has the highest similarity. The optimal bipartite matching $\Pi^{*}$ can be calculated efficiently by the Hungarian algorithm~\cite{kuhn1955hungarian}.   

Once the matched pairs of $S$ and $\tilde{S}$ are established, the proposed loss function tailored for 3D object detection can be computed as:
\begin{equation} \label{eq_loss}
\begin{aligned}
    \mathcal{L}_{det} =& \sum_{(i \rightarrow j) \in \Pi^*} [ \mathcal{L}_{cls}(W^C_i, \tilde{W}^C_j) +\\ &\mathbbm{1}_{\{\tilde{S}_i \neq \varnothing\}}\lambda_{reg} \mathcal{L}_{reg}(W^{\{R, L, O, S\}}_i, \tilde{W}^{\{R, L, O, S\}}_j)],
\end{aligned}
\end{equation}
where $\mathcal{L}_{cls}$ is the focal loss applied on the predicted and target category words, and $\mathcal{L}_{reg}$ takes the words $W^{\{R, L, O, S\}}$ and $\tilde{W}^{\{R, L, O, S\}}$ as input and translates them back to the respective object's location and shape $[x, y, z, l, w, h, \theta]$ on which the smooth-$L_{1}$ loss is then applied. The indicator function $\mathbbm{1}_{\{\tilde{S}_i \neq \varnothing\}}$ implies that we only apply $\mathcal{L}_{reg}$ on those sequences that are matched to the ground truths, and $\lambda_{reg}$ is a coefficient that balances the two losses.

Since each ground truth 3D object is matched with only one predicted sequence, the scene-to-sequence head does not produce duplicated boxes for an individual object. Hence the time-consuming process of non-maximum suppression can be eliminated in our framework. During the inference stage, we simply filter out those low-quality sequences in which the maximum class probability in $W^{C}$ is below a certain threshold, and we translate the remaining sequences into 3D objects as the final detection results.  

\subsection{Discussion} \label{3.4}
Our proposed Point2Seq shares a similar intuition with the concurrent work Pix2Seq~\cite{chen2021pix2seq}, which is proposed for image-based object detection, in terms of leveraging objects as words that can be read out from a feature map. However, our method is intrinsically different from~\cite{chen2021pix2seq} in $3$ aspects: 1) Unlike~\cite{chen2021pix2seq} that merges all objects into an individual sequence, we treat each object as a sequence and predict all objects in parallel, while words in each object are generated sequentially. In this manner, we can circumvent the object ordering problem in~\cite{chen2021pix2seq}, and our method is much more efficient at the inference stage compared to~\cite{chen2021pix2seq} in which the inference latency will be heavily influenced by the total object count in an image. 2) We adopt the continuous word representations instead of discrete tokens in~\cite{chen2021pix2seq}. The use of continuous representations relieves the need for quantization and makes our method compatible with the existing loss functions tailored for 3D object detection. 3) We propose the scene-to-sequence decoder to generate words for each object, in lieu of the Transformer architecture in~\cite{chen2021pix2seq}. The scene-to-sequence decoder is lightweight and leverages a sparse set of features to predict each object, which is more suitable for 3D object detection where the detected targets usually are small and sparse.

\begin{table*}[htbp]
\centering
\setlength{\tabcolsep}{3.7mm}{
\small
\begin{tabular}{l|c|c|cc|cc}
\toprule
\multirow{2}{*}{Method} & \multirow{2}{*}{Backbone} & \multirow{2}{*}{Head} & \multicolumn{2}{c|}{Vehicle LEVEL 1} & \multicolumn{2}{c}{Vehicle LEVEL 2} \\ 
&    &    & 3D mAP(\%) & 3D mAPH(\%) & 3D mAP(\%) & 3D mAPH(\%) \\ \midrule
LaserNet~\cite{Meyer2019LaserNetAE}        & Range           & Anchor            & 52.1           & 50.1           & -             & -     \\
RCD~\cite{bewley2020range}                 & Range           & Center            & 69.0           & 68.5
        & -             & -     \\
RangeDet~\cite{fan2021rangedet}            & Range           & Center            & 72.85          & - 
        & -             & -     \\
RSN~\cite{sun2021rsn}                      & Range           & Center            & 75.1           & 74.6
        & 66.0          & 65.6  \\
PointPillars~\cite{lang2019pointpillars}   & Pillar          & Anchor            & 63.3           & 62.7           & 55.2          & 54.7  \\
Pillar-OD~\cite{wang2020pillar}            & Pillar          & Anchor            & 69.8           & -              & -             & -     \\
MVF~\cite{zhou2020end}                     & Voxel           & Anchor            & 62.93          & -
        & -             & -     \\
PV-RCNN~\cite{shi2020pv}                   & Voxel           & Anchor            & 77.51          & 76.89          & 68.98         & 68.41 \\
VoTr-TSD~\cite{mao2021voxel}               & Voxel           & Anchor            & 74.95          & 74.25          & 65.91         & 65.29 \\
Voxel R-CNN~\cite{deng2020voxel}           & Voxel           & Anchor            & 75.59          & -              & 66.59         & -     \\
Pyramid-RCNN~\cite{mao2021pyramid}         & Voxel           & Anchor            & 76.3           & 75.68          & 67.23         & 66.68 \\
CT3D~\cite{Sheng_2021_ICCV}                & Voxel           & Anchor            & 76.3           & -              & 69.04         & -    \\
CVCNet~\cite{chen2020every}                & Voxel           & Center            & 65.20          & -
        & -             & -    \\
AFDet~\cite{ge2020afdet}                   & Voxel           & Center            & 63.69          & -
        & -             & -     \\ 
CenterPoints~\cite{yin2021center}          & Voxel           & Center            & 76.7           & 76.2           & 68.8          & 68.3  \\ \midrule
SECOND$^{\dag}$~\cite{yan2018second}      & Voxel           & Anchor            & 73.62          & 73.14          & 64.86         & 64.40   \\
CenterPoints$^{\dag}$~\cite{yin2021center} & Voxel           & Center            
& 75.58
& 75.01 
& 67.00 
& 66.52 \\
Point2Seq (Ours)                         & \textbf{Voxel}  & \textbf{Sequence} 
& \textbf{77.52}
& \textbf{77.03} 
& \textbf{68.80} 
& \textbf{68.36}\\
\bottomrule
\end{tabular}}
\setlength{\belowcaptionskip}{10pt}
\vspace{-0.5em}
\caption{Performance comparison on the Waymo Open Dataset with 202 validation sequences for vehicle detection. $\dag$: re-implemented using the official code. Point2Seq maintains the same backbone, data augmentations, and training epochs with the re-implemented baselines.} \label{table_waymo_val}
\vspace{-4mm}
\end{table*}

\begin{table*}[]
\centering
\setlength{\tabcolsep}{1.0mm}{
\small
\begin{tabular}{c|c|cccc|cccc|cccc}
\toprule
\multicolumn{1}{c|}{\multirow{2}{*}{Method}} & \multicolumn{1}{c|}{\multirow{2}{*}{mAP(\%)}} 
& \multicolumn{4}{c|}{Vehicle mAP(\%)} & \multicolumn{4}{c|}{Pedestrian mAP(\%)} & \multicolumn{4}{c}{Cyclist mAP(\%)} \\
\multicolumn{1}{c|}{} 
& \multicolumn{1}{c|}{} & overall & 0-30m & 30-50m 
& \multicolumn{1}{c|}{50m-inf} & overall & 0-30m & 30-50m 
& \multicolumn{1}{c|}{50m-inf} & overall & 0-30m & 30-50m & 50m-inf \\ \midrule
PointRCNN~\cite{shi2020pv}               & 28.74 & 52.09 & 74.45 & 40.89 & 16.81 &  4.28 &  6.17 &  2.4  &  0.91 & 29.84 & 46.03 & 20.94 &  5.46 \\
PointPillars~\cite{lang2019pointpillars} & 44.34 & 68.57 & 80.86 & 62.07 & 47.04 & 17.63 & 19.74 & 15.15 & 10.23 & 46.81 & 58.33 & 40.32 & 25.86 \\
PV-RCNN~\cite{shi2020pv}                 & 53.55 & 77.77 & 89.39 & 72.55 & 58.64 & 23.50 & 25.61 & 22.84 & 17.27 & 59.37 & 71.66 & 52.58 & 36.17 \\ \midrule
SECOND~\cite{yan2018second}              & 51.89 & 71.19 & 84.04 & 63.02 & 47.25 & 26.44 & 29.33 & 24.05 & 18.05 & 58.04 & 69.96 & 52.43 & 34.61 \\
CenterPoints~\cite{yin2021center}        & 60.05 & 66.79 & 80.10 & 59.55 & 43.39 & 49.90 & 56.24 & 42.61 & 26.27 & 63.45 & 74.28 & 57.94 & 41.48 \\
Point2Seq (Ours)                       & \textbf{66.16} 
& \textbf{73.43} & \textbf{85.16} & \textbf{66.21} & \textbf{50.76}
& \textbf{57.53} & \textbf{68.21} & \textbf{47.15} & \textbf{25.18}
& \textbf{67.53} & \textbf{77.95} & \textbf{62.14} & \textbf{46.06}
\\ \bottomrule
\end{tabular}}
\vspace{-0.5em}
\caption{Performance comparison on the ONCE dataset validation split. Point2Seq maintains the same backbone architecture and training configurations with the baselines on the ONCE benchmark.} \label{table_once_val}
\vspace{-5mm}
\end{table*}

\section{Experiment}
In this section, we evaluate Point2Seq on the commonly-used Waymo Open Dataset~\cite{sun2020scalability} and the ONCE dataset~\cite{mao2021one}. We first introduce the experimental settings in Sec.~\ref{4.1}. Then we compare our approach with previous state-of-the-art methods on the Waymo Open Dataset (Sec.~\ref{4.2}) and the ONCE dataset (Sec.~\ref{4.3}). Finally, we report the inference speed and the number of parameters, as well as the efficacy of different components in our model in Sec.~\ref{4.4}. 

\subsection{Experimental Setup} \label{4.1}
\textbf{Waymo Open Dataset.} The Waymo Open Dataset is composed of $1000$ sequences of point clouds, in which $798$ sequences (nearly $158k$ point cloud samples) are used as the training set, and $202$ sequences (nearly $40k$ point cloud samples) are utilized as the validation set. The evaluation metrics on the Waymo Open Dataset are 3D mean Average Precision (mAP) and mAP weighted by heading accuracy (mAPH). The IoU threshold used for vehicles is $0.7$ and $0.5$ for other categories. The detection results are reported based on the difficulty levels: LEVEL 1 for boxes with more than $5$ points and LEVEL 2 for boxes with at least $1$ point.

\textbf{ONCE Dataset.} The ONCE dataset contains one million point clouds in total, in which $5$k, $3$k, $8$k point clouds are annotated as the training, validation, testing split, respectively. The remaining point clouds are kept unannotated for self-/semi-supervised learning. In this paper, we train our model on the training split and report the detection results of vehicles, pedestrians, and cyclists on the validation and testing split, without using the unlabeled data. The official evaluation metric is mean Average Precision (mAP), and the detection results are divided according to the objects' distances to the sensor: $0$-$30$m, $30$-$50$m, and $50$m-inf.

\textbf{Implementation Details.} On the Waymo Open Dataset, we use the same 3D sparse convolutional neural network and 2D convolutional neural network as~\cite{yin2021center}. The input voxel size and the output resolution of the BEV feature map are also kept the same as~\cite{yin2021center} for a fair comparison. On the ONCE dataset, all the voxel-based detectors use the same type of 3D backbone~\cite{yan2018second} in their official benchmark implementations. We also follow the setting and use the 3D backbone in~\cite{yan2018second}. For other model configurations, we adopt the same as those on the ONCE benchmark.  

\textbf{Training and Inference Details.} We train our model with the ADAM optimizer and the cosine annealing learning rate scheduler. On the Waymo Open Dataset, we uniformly sample $20\%$ of the point cloud samples for training and use the full validation set for evaluation following~\cite{shi2020pv}. We train our model with the batch size $32$ and the initial learning rate $0.006$ for $180$ epochs on $8$ V100 GPUs. $\lambda_{reg}$ in the loss function is set to $2$. Data augmentations are kept the same as~\cite{yin2021center}. On the ONCE dataset, we follow the training settings of the respective benchmark and train our model with the batch size $32$ and the initial learning rate $0.003$ for $80$ epochs on $8$ V100 GPUs. $\lambda_{reg}$ in the loss function is set to $0.5$. Data augmentations are kept the same as~\cite{mao2021one}. On both two datasets, we filter out those objects with the maximum foreground class probability in $W^{C}$ below $0.2$ and keep the remaining objects as the final detection results during the inference stage, without  any other post-processing.

\subsection{Comparisons on the Waymo Open Dataset} \label{4.2}
Since our contribution focuses on the 3D object prediction mechanism, the fairest way to evaluate our method and compare it with the anchor-based and center-based methods is to only replace the center or anchor head with our scene-to-sequence decoder while maintaining other components as the same. We follow this principle and re-implement two baseline models from their official implementations: SECOND~\cite{yan2018second} with the anchor head and CenterPoints~\cite{yin2021center} with the center head. Our proposed Point2Seq, re-implemented SECOND, and CenterPoints have the same voxel-based 3D backbone, data augmentations, and training epochs to ensure a completely fair comparison.

Table~\ref{table_waymo_val} shows the detection results on the Waymo validation set. Simply switching from the anchor and center head to our Point2Seq gives $3.90\%$ and $1.94\%$ LEVEL 1 mAP improvements, respectively. Our method attains $77.52\%$ LEVEL 1 mAP and $68.80\%$ LEVEL 2 mAP for vehicle detection, surpassing existing methods by a significant margin. Our approach even outperforms those time-consuming two-stage 3D detectors~\cite{shi2020pv, mao2021pyramid, Sheng_2021_ICCV}, which indicates the effectiveness of the scene-to-sequence decoder.

\subsection{Comparisons on the ONCE dataset} \label{4.3}
The ONCE dataset benchmarks different voxel-based detectors using the same backbone network, and we also follow this rule for a fair comparison. As is shown in Table~\ref{table_once_val}, Point2Seq attains the state-of-the-art results on all classes, with $73.43\%$ mAP for vehicle detection, $57.53\%$ mAP for pedestrian detection, and $67.53\%$ for cyclist detection. The overall mAP of our approach is $66.16\%$, $6.11\%$ higher than the center-based 3D object detector~\cite{yin2021center} and $14.27\%$ higher than the anchor-based 3D object detector~\cite{yan2018second}. The observations on the ONCE dataset are consistent with those on the Waymo Open Dataset.

\subsection{Ablation Studies} \label{4.4}

\textbf{Inference speed and model parameters.} Table~\ref{table_abl_eff} demonstrates the inference time and the number of parameters of our method. Since the 3D objects in a scene are predicted in parallel, Point2Seq can obtain high efficiency with $70.4$ms inference latency on average for a single model on a V100 GPU. The scene-to-sequence head only contains several linear projection layers, and the sampling operation is parameter-free, so the model only introduces $0.1$M additional parameters compared to the center-based baseline.   

\textbf{Generalizability on different 3D backbones.} To verify whether Point2Seq can achieve superior performances upon different backbones, we apply the scene-to-sequence head on both the voxel-based~\cite{yin2021center} and pillar-based~\cite{lang2019pointpillars} 3D backbones and compare the results with the center and anchor head, respectively. Table~\ref{table_abl_head} demonstrates that on both two types of backbone networks, our method consistently outperforms the anchor-based and center-based detectors.  

\textbf{Effects of different components in Point2Seq.} Table~\ref{table_abl_component} shows the effectiveness of the scene-to-sequence decoder and the similarity-based sequence matching scheme. Similarity-based sequence matching can be independently applied on the previously used convolutional head and boost the detection performance by $3.1\%$ mAP compared to the anchor-based baseline. Combing the two proposed components, we can obtain a performance gain of $3.9\%$ mAP.

\begin{table}[]
\small
\centering
\setlength{\tabcolsep}{1.55mm}{
\begin{tabular}{c|c|c|c}
\toprule
\multirow{2}{*}{Method} & Vehicle        & \multirow{2}{*}{\#Param} & \multirow{2}{*}{Latency (ms)} \\ 
                         & L1/L2 mAP(\%) &                                   &                               \\ \midrule
PV-RCNN~\cite{shi2020pv}                    & 77.51/68.98 & 13.05M & 300    \\ 
CenterPoints~\cite{yin2021center}           & 76.7/68.8   &  8.74M &  77    \\ \midrule 
SECOND$^{\dag}$~\cite{yan2018second}        & 73.62/64.86 &  7.28M &  66.5  \\
CenterPoints$^{\dag}$~\cite{yin2021center}  & 75.58/67.00 &  7.76M &  69.5  \\
Point2Seq$^{\dag}$ (Ours)                   & 77.52/68.80 &  7.86M &  70.4  \\ \bottomrule                      
\end{tabular}}
\vspace{-0.7em}
\caption{Inference speed and parameters amount. $\dag$: tested under the same environment using a single model on a V100 GPU.} 
\label{table_abl_eff}
\vspace{-1mm}
\end{table}

\begin{table}[]
\centering
\setlength{\tabcolsep}{1.8mm}{
\small
\begin{tabular}{c|c|c|c}
\toprule
\multicolumn{1}{c|}{\multirow{2}{*}{Backbone}} & \multicolumn{1}{c|}{\multirow{2}{*}{Head}} & \multicolumn{1}{c|}{Veh. LEVEL 1} & Veh. LEVEL 2      \\  
\multicolumn{1}{c|}{}                          & \multicolumn{1}{c|}{}                      & \multicolumn{1}{c|}{mAP/mAPH(\%)}    & mAP/mAPH(\%)     \\ \midrule
\multirow{3}{*}{Pillar}                       
& Anchor          & 63.31/62.74          & 55.24/54.72          \\
& Center          & 65.46/64.66          & 57.59/56.88          \\
& Point2Seq           & \textbf{69.01/68.25} & \textbf{60.72/60.03} \\ \midrule
\multirow{3}{*}{Voxel}                        
& Anchor          & 73.62/73.14          & 64.86/64.40           \\
& Center          & 75.58/75.01          & 67.00/66.52          \\
& Point2Seq           & \textbf{77.52/77.03} & \textbf{68.80/68.36} \\ \bottomrule
\end{tabular}}
\setlength{\belowcaptionskip}{10pt}
\vspace{-0.7em}
\caption{Performances on different backbone networks.} 
\label{table_abl_head}
\vspace{-8mm}
\end{table}

\textbf{The order of words in a sequence.} We explored the influence of changing the predicted word orders in our method. The results in Table~\ref{table_abl_order} indicate that the order of words plays a non-negligible role on the detection performance. 
For example, putting $W^{O}$ at the $4$th place  greatly reduces the detection accuracy, which may indicate the importance of predicting the object orientation at an earlier position. Putting $W^{C}$ at the end will exhibit better performance compared to putting $W^{C}$ at the beginning. 

\textbf{The choice of different similarity metrics.} We evaluated different formulas of the similarity metric used in similarity-based sequence matching. Table~\ref{table_abl_metric} exhibits the results of the $3$ formulas we have examined. The formula (3) is the currently adopted similarity metric. The formula (2) replaces the term $e^{-(1-\alpha) \sum_{j \in \{R, O, S, L\}} |W^{j} - \tilde{W}^{j}|}$ in Eq.~\ref{eq_sim} with $e^{-(1-\alpha) 3DIoU(B, \tilde{B}) }$ where $3DIoU(B, \tilde{B})$ computes the 3D IoU score of two bounding boxes. The formula (1) replaces the same term with $e^{-(1-\alpha)\sum_{j = 1}^{8} |C^{j} - \tilde{C}^{j}|}$, where we calculate the differences of $8$ respective corners $C$ of two bounding boxes. The results indicate that Eq.~\ref{eq_sim} is the best among the $3$ formulas as the similarity metric.

\section{Conclusion}
We present Point2Seq, an effective and general 3D object detection framework that can be applied to most grid-based backbone networks. Point2Seq contains a scene-to-sequence decoder, which can auto-regressively generate sequences describing the detected 3D objects, and similarity-based sequence matching is proposed to enable end-to-end training without human-designed label assignments. For future works, we plan to extend our framework to multi-modality 3D object detection.

\begin{table}[]
\setlength{\tabcolsep}{1.9mm}{
\small
\centering
\begin{tabular}{c|c|c|c}
\toprule
\multicolumn{1}{c|}{\multirow{2}{*}{Assignment}} & \multicolumn{1}{c|}{\multirow{2}{*}{Module}} 
& \multicolumn{1}{c|}{Veh. LEVEL 1} & \multicolumn{1}{c}{Veh. LEVEL 2} \\ 
\multicolumn{1}{c|}{} & \multicolumn{1}{l|}{} & \multicolumn{1}{c|}{mAP/mAPH(\%)} & \multicolumn{1}{c}{mAP/mAPH(\%)} \\ \midrule
Anchor     & C.H    & 73.62/73.14  & 64.86/64.40 \\
Center     & C.H    & 75.58/75.01  & 67.00/66.52 \\
S.S.M.     & C.H    & 76.72/76.19  & 68.02/67.54 \\
S.S.M.     & S.S.D  & \textbf{77.52/77.03}  & \textbf{68.80/68.36} \\ \bottomrule                     
\end{tabular}}
\setlength{\belowcaptionskip}{10pt}
\vspace{-0.7em}
\caption{Effects of different components in Point2Seq. C.H: Convolutional Head in previous works. S.S.M.: Similarity-based Sequence Matching. S.S.D: Scene-to-Sequence Decoder.} 
\label{table_abl_component}
\vspace{-4mm}
\end{table}

\begin{table}[]
\small
\centering
\setlength{\tabcolsep}{1mm}{
\begin{tabular}{c|c|c}
\toprule
\multicolumn{1}{c|}{\multirow{2}{*}{Order}} & \multicolumn{1}{c|}{Veh. LEVEL 1} & Veh. LEVEL 2 \\ 
\multicolumn{1}{c|}{}                       & \multicolumn{1}{c|}{mAP/mAPH(\%)} & mAP/mAPH(\%) \\ \midrule
$W^{R}, W^{O}, W^{S}, W^{L}, W^{C}$         & 77.40/76.91                       & 68.71/68.27  \\
$W^{R}, W^{O}, W^{L}, W^{S}, W^{C}$         & 77.43/76.94                       & 68.74/68.30  \\
$W^{R}, W^{L}, W^{O}, W^{S}, W^{C}$         & 77.52/77.03                       & 68.80/68.36  \\
$W^{R}, W^{L}, W^{S}, W^{O}, W^{C}$         & 73.82/73.24                       & 66.16/65.61  \\
$W^{R}, W^{C}, W^{L}, W^{O}, W^{S}$         & 75.96/75.41                       & 67.26/66.76  \\
$W^{R}, W^{C}, W^{S}, W^{O}, W^{L}$         & 74.52/73.99                       & 66.86/66.35  \\
$W^{R}, W^{C}, W^{O}, W^{L}, W^{S}$         & 77.05/76.54                       & 68.37/67.90  \\
$W^{R}, W^{C}, W^{O}, W^{S}, W^{L}$         & 76.82/76.28                       & 68.11/67.62  \\
\bottomrule
\end{tabular}}
\setlength{\belowcaptionskip}{10pt}
\vspace{-0.7em}
\caption{Effects of different word orders in the sequences.} 
\label{table_abl_order}
\vspace{-4mm}
\end{table}

\begin{table}[]
\setlength{\tabcolsep}{4.9mm}{
\small
\centering
\begin{tabular}{c|c|c}
\toprule
\multicolumn{1}{c|}{\multirow{2}{*}{Metrics}} & \multicolumn{1}{c|}{Veh. LEVEL 1} & Veh. LEVEL 2 \\ 
\multicolumn{1}{c|}{}                         & \multicolumn{1}{c|}{mAP/mAPH(\%)} & mAP/mAPH(\%) \\ \midrule
(1)                                           & 74.31/73.52                       & 65.61/65.04  \\
(2)                                           & 75.07/74.43                       & 66.36/65.73  \\
(3)                                           & \textbf{75.40/74.81}                        & \textbf{66.76/66.23} \\ \bottomrule
\end{tabular}}
\setlength{\belowcaptionskip}{10pt}
\vspace{-0.7em}
\caption{Comparisons of different similarity metrics. } 
\label{table_abl_metric}
\vspace{-8mm}
\end{table}

\section*{Acknowledgement} This work is supported by NUS Faculty Research Committee Grant (WBS: A-0009440-00-00), and NRF Centre for Advanced Robotics Technology Innovation (CARTIN) Project.

\clearpage
{\small
\bibliographystyle{ieee_fullname}
\bibliography{egbib}
}

\end{document}